\documentclass[twocolumn]{svjour3}          

\usepackage{amsthm}
\usepackage{graphicx}
\usepackage{amssymb}
\usepackage{color}
\usepackage{amsmath,amsfonts}

\usepackage{algorithm}
\usepackage{array}

\usepackage{subfigure}

\usepackage{bbm}
\usepackage{subcaption}
\usepackage[english]{babel}
\usepackage[utf8]{inputenc} 
\usepackage[T1]{fontenc}    
\usepackage{hyperref}       
\usepackage{url}            
\usepackage{booktabs}       
\usepackage{nicefrac}       
\usepackage{microtype}      
\usepackage{makecell}
\usepackage{colortbl} 
\usepackage{xcolor}
\definecolor{mygray}{gray}{.1}
\definecolor{mygray2}{gray}{.9}
\definecolor{mygray}{gray}{.9}
\usepackage{amsthm}
\usepackage{multirow}
\usepackage{mathrsfs}
\usepackage{float}
\usepackage[misc]{ifsym}
\usepackage{stfloats}
\usepackage{caption}
\usepackage{algorithm}
\usepackage{algpseudocode} 

\usepackage{pythonhighlight}
\usepackage{bm}
\usepackage{amsmath}               
  {
      \theoremstyle{plain}
      
  }

\journalname{The Visual Computer} 

\begin{document}

\title{SemanticStitch: Enhancing Image Coherence through Foreground-Aware Seam Carving}

\subtitle{}

\author{Ji-Ping Jin$^*$ \and Chen-Bin Feng$^*$ \and Rui Fan \and Chi-Man Vong}

\institute{
Ji-Ping Jin ; Rui Fan \at ShanghaiTech University, Shanghai, China \\
Chen-Bin Feng ; Chi-Man Vong \at University of Macau, Macau, China \\
$^*$These authors contributed equally to this work. \\
Corresponding author: Rui Fan ; Chi-Man Vong  \\ E-mail: cmvong@um.edu.mo ; Fanrui@shanghaitech.edu.cn \\
}
\date{ }

\maketitle
\begin{figure}[t]
    \centering
    \includegraphics[width=1\linewidth]{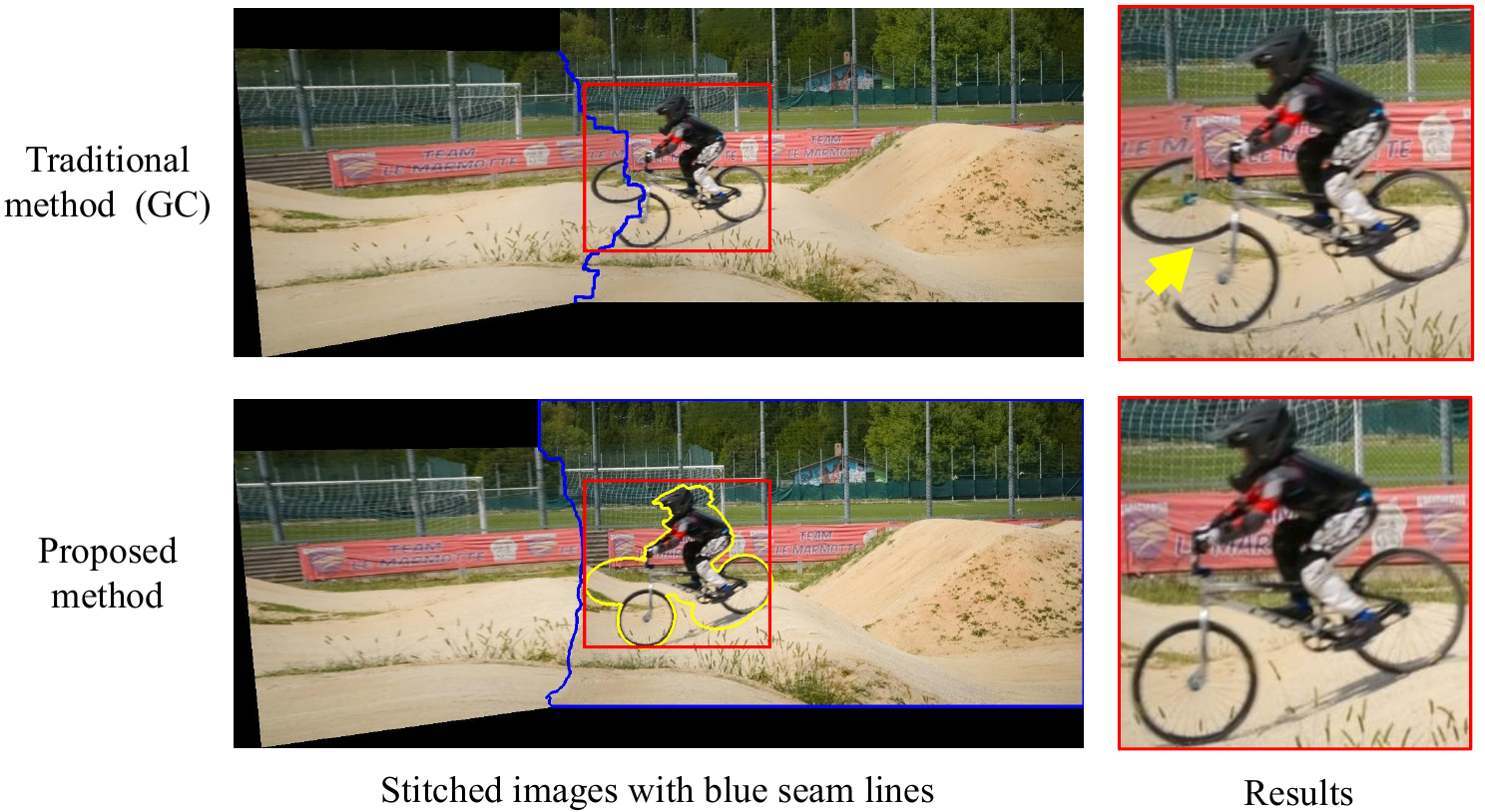}
    \caption{\textbf{Comparison of traditional methods (e.g., Graph Cut)} Fair comparisons across different datasets and methods are provided in subsequent sections.}
    \label{fig:Illustration_1}
\end{figure}
\begin{abstract}
    Image stitching often faces challenges due to varying capture angles, positional differences, and object movements, leading to misalignments and visual discrepancies. Traditional seam carving methods neglect semantic information, causing disruptions in foreground continuity. We introduce SemanticStitch, a deep learning-based framework that incorporates semantic priors of foreground objects to preserve their integrity and enhance visual coherence. Our approach includes a novel loss function that emphasizes the semantic integrity of salient objects, significantly improving stitching quality. We also present two specialized real-world datasets to evaluate our method's effectiveness. Experimental results demonstrate substantial improvements over traditional techniques, providing robust support for practical applications. The codes are available at  \url{https://github.com/Pokerman8/OAIV-Coherence}.

\end{abstract}

\section{Introduction}
In the field of image stitching, disparities in capture angles, positional differences, and movements of objects within the scene often result in significant misalignments and visual discrepancies between the images to be stitched.

Historically, approaches such as Graph Cut\cite{kwatra2003graphcut} and Dynamic Programming\cite{duplaquet1998building} have been employed to address these issues. However, these methods have not adequately considered the semantic information of the images, often resulting in stitching lines that traverse foreground objects. This causes significant discontinuities and mismatches in the visual attributes of the foreground objects on either side of the stitching line.

To address these challenges, we propose an innovative deep learning-based image stitching framework. Our method leverages semantic priors of foreground objects, incorporating their semantic attributes to avoid stitching lines crossing critical foreground objects.  By incorporating these semantic attributes, our framework ensures that the foreground objects are preserved and seamlessly integrated across the stitched image.This design enhances the visual coherence and overall aesthetic of the stitched image while maintaining smooth transitions at the seams.

Similar to introducing boundary-aware mechanisms in segmentation tasks~\cite{tang2024boundary} to enhance object integrity, our method incorporates semantic constraints in image stitching to ensure that the seams do not disrupt the boundaries of foreground objects.

We introduce a novel loss function based on salient object semantic integrity to improve existing deep learning-based image stitching methods. Our approach addresses the limitations of traditional seam-cutting and feature-based methods, emphasizing the preservation of semantic information of salient objects. This significantly enhances the overall quality and realism of the stitched images. 

The new loss function ensures the seamless integration and preservation of important objects within the image, improving alignment and visual consistency. By combining powerful feature extraction with advanced semantic analysis, we achieve superior image stitching results.

\begin{figure*}[!t]
    \centering
    \includegraphics[width=\textwidth]{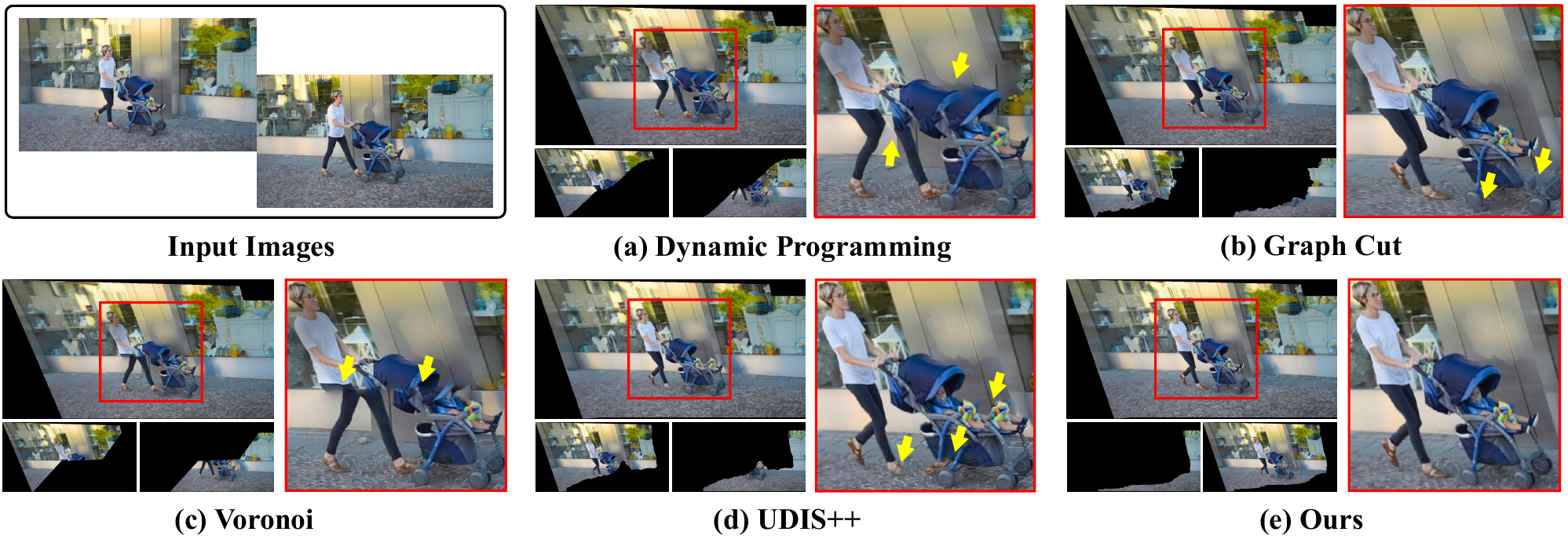}%
    \caption{\textbf{Comparison of image stitching methods. } This figure illustrates the performance of our approach relative to other mainstream seam-based methods. The magnified views show that our method significantly outperforms others due to its object-aware design. Yellow arrows indicate foreground objects that are incorrectly truncated by the other methods. The two images in the bottom left corner of each method depict the seam carving process, which are then stitched together to produce the final result.}
    \label{fig:teaser}
\end{figure*}

Given the unique nature of image stitching, where stitching lines often intersect objects, traditional image stitching datasets do not meet specific requirements.To better accommodate this task, we have compiled and constructed specialized real-world test datasets, including a dataset derived from processing DAVIS\cite{perazzi2016benchmark}, designed to test scenarios involving moving foreground objects. These datasets cover various complex scenes where stitching lines intersect objects, supporting the testing and evaluation of deep learning models for this task. This initiative aims to enhance the generalization and practicality of the models, ensuring more accurate and natural stitching results in real-world applications.

Experimental results demonstrate that our method significantly improves the quality of stitched images, effectively addressing the issues present in traditional techniques. This research not only enriches the theoretical foundations of the image stitching field but also provides robust technical support for related applications.

Our contributions can be summarized as follows:
\begin{itemize}
    \item We introduce an object-aware seam carving framework that includes a saliency-driven network and saliency-aware seam carving loss.
    \item We propose two specialized real-world datasets for testing and evaluating the integrity of foreground objects in stitched images.
    \item We design an advanced network architecture tailored for seam carving, which is an improvement over previous networks in terms of performance and efficiency.
\end{itemize}

\section{Related Work}

In this section, we review existing methodologies relevant to our proposed approach in the domain of computer vision. Our discussion is bifurcated into two primary categories: traditional methods based on computer graphics and contemporary methods leveraging deep learning techniques. 

A significant focus is given to the overlap regions in image stitching, as these are critical in determining the overall aesthetic and functional success of the composite images. The human eye tends to focus on salient foreground objects within these overlap areas. Consequently, the effectiveness with which these salient objects are blended plays a pivotal role in the perceived quality of the final stitched image.

\subsection{Traditional Image Stitching}
In traditional image stitching techniques, seam cutting is prominently utilized. This approach employs graph-cut optimization to minimize various energy functions, effectively transforming seam prediction into a classical minimal cut problem, thereby yielding reasonable stitching outcomes \cite{Kwatra_Schödl_Essa_Turk_Bobick_2003}. Alternatively, methods based on the color differences in image overlap regions construct a cost graph using Dynamic Programming (DP). The optimal path in this graph is then determined through dynamic programming algorithms aimed at finding a local optimum \cite{duplaquet1998building}.

However, these techniques predominantly focus on minimizing gradient differences \cite{Dai_Fang_Li_Zhang_Zhou_2021}, Euclidean metric color differences \cite{Kwatra_Schödl_Essa_Turk_Bobick_2003}, and motion- and exposure-aware differences \cite{Eden_Uyttendaele_Szeliski_2006}. They often overlook the semantic consistency of foreground objects in the image. This oversight can result in seams that intersect foreground objects, leading to poor quality seams and, consequently, unsatisfactory stitching results.

\begin{figure*}[htbp]
    \centering
    \includegraphics[width=\textwidth]{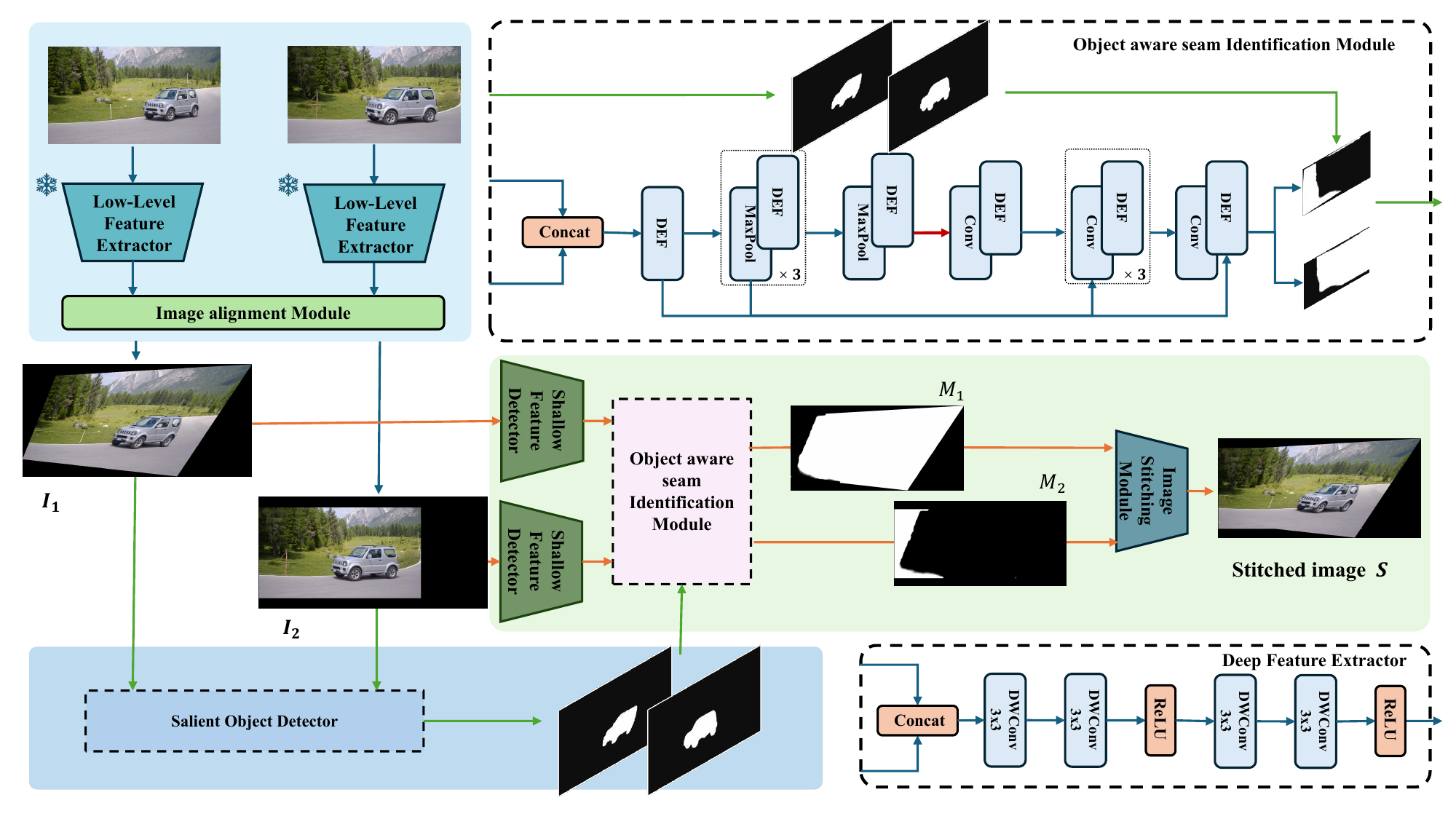}
    \caption{\textbf{Overview of the proposed method. } The first component predicts a reliable warp between two images. The second component predicts the foreground object masks for both images. The third component determines a seam that preserves the integrity of the foreground objects during stitching.}
    \label{fig:model}
\end{figure*}

\subsection{Deep Learning-Based Image Stitching}
In recent years, deep learning methods have significantly advanced, with techniques that utilize supervised learning \cite{Nie_Lin_Liao_Zhao_2022,Song_Um_Lee_Cho_2021} and weak supervision \cite{Song_Lee_Lee_Um_Cho_2022} to automatically extract high-level semantic features from extensive datasets. Methods based on deep learning for multi-scale feature extraction, such as DDMSNet\cite{zhang2021deep}, DBLRNet\cite{zhang2018adversarial}, and Gridformer\cite{li2024gridformer}, have made feature extraction increasingly accurate.These methods have proven robust across various challenging scenarios.

Deep learning has also advanced multispectral stitching to handle spectral, illumination, and parallax challenges. \cite{jiang2024multispectral} uses global-aware quadrature pyramids for robust alignment, \cite{jiang2022towards} introduces a progressive pyramid and the MSIS dataset, and \cite{jiang2023multi} applies spatial graph reasoning for seamless fusion. These works demonstrate deep learning's potential for multispectral image stitching.

Nie et al. \cite{Nie_Lin_Liao_Liu_Zhao_2021} propose an innovative unsupervised deep learning framework for image stitching that circumvents the limitations associated with feature-based and supervised methods. Their approach involves a two-stage process: initial unsupervised coarse alignment followed by feature-to-pixel image reconstruction, enhancing the adaptability and accuracy of the stitching process.

Recent advancements have also seen the emergence of deep seam stitching algorithms such as UDIS++ \cite{Nie_Lin_Liao_Liu_Zhao_2023} and DSP \cite{cheng2023deep}. These methods employ soft encoding techniques during the seam mask generation process to facilitate backpropagation, which is typically hindered by binary masks. Despite their advantages, soft encoding often struggles to delineate strict seam boundaries clearly, posing challenges in maintaining precise alignment.

In reference-based super-resolution(RefSR), a similar challenge exists in that not all information from the reference image is suitable for direct use; thus, an attention-based complementary information fusion strategy has been proposed~\cite{wang2024high}. Inspired by this, our method also selectively leverages semantic priors to guide seam placement.

\section{Method}

\subsection{Overview}
The architecture of the proposed method, as illustrated in Figure~\ref{fig:model}, is divided into three core components. The first component addresses the warp problem between the two images to be stitched. The second component derives reliable and accurate masks for the foreground objects. The third component produces an image stitching result that strives to preserve the integrity of the foreground objects as much as possible.

In the image alignment phase, we use ResNet50\cite{he2016deep} for multi-scale feature extraction\cite{jin2025mb,wang2013estinet}, followed by progressive regression and contextual correlation to predict and refine the 4-point homography and flexible warping transformations\cite{Nie_Lin_Liao_Liu_Zhao_2023}.

In the salient object detection phase, we employ a method utilizing a Transformer-based network\cite{yun2022selfreformer} to integrate both global and local context information. This approach employs a Pyramid Vision Transformer  as the encoder backbone, which effectively captures long-range dependencies and preserves the integrity of foreground object detection. Additionally, a two-stage Context Refinement Module  is used to fuse global and local contexts, thereby refining prediction details with high accuracy.

In the image fusion stage, utilizing the foreground object integrity information derived from the second stage's salient object detector, images are stitched using soft-coded seam reconstruction with feature differentials. A shallow feature extractor scales image features, which are then processed by a UNET-structured network\cite{ronneberger2015u}. An upsampling-based seam generator forecasts the seam mask, and weighted fusion produces the final stitched image\cite{mei2024dunhuangstitch}.

In detail, given two misaligned target images $I_t \in \mathbb{R}^{3 \times H \times W}$ and reference images $I_r \in \mathbb{R}^{3 \times H \times W}$, the Image Alignment Module processes these inputs to produce warped images $I_{wt} \in \mathbb{R}^{3 \times H_s \times W_s}$ and $I_{wr} \in \mathbb{R}^{3 \times H_s \times W_s}$. Subsequently, the Salient Object Detector is employed to generate the corresponding foreground object masks $M_t \in \mathbb{R}^{1 \times H_s \times W_s}$ and $M_r \in \mathbb{R}^{1 \times H_s \times W_s}$, which are then combined to form $M_{object} = M_t \cap M_r$. The warped images $I_{wt}$ and $I_{wr}$ are processed through identical shallow feature detectors to produce feature maps $F_{wt} \in \mathbb{R}^{3 \times H_s \times W_s}$ and $F_{wr} \in \mathbb{R}^{3 \times H_s \times W_s}$. In the Object Aware Seam Identification Module, these feature maps are refined using a U-net-like network to generate two masks, $M_{learn_1}$ and $M_{learn_2}$, as well as the corresponding seam lines. By calculating the proposed loss function with $M_{object}$, the optimization process is conducted, ultimately resulting in a seam line that ensures the integrity of the foreground objects in the stitched image.

\subsection{Salient Object Detector}

In our approach, we utilize a salient object detector based on the method described in SelfReformer \cite{yun2022selfreformer}. This method employs a Transformer-based network to effectively integrate both global and local context information. Specifically, a Pyramid Vision Transformer is used as the encoder backbone, which excels at capturing long-range dependencies and preserving the integrity of foreground object detection. Additionally, a two-stage Context Refinement Module (CRM) fuses global and local contexts, thereby refining prediction details with high accuracy.

For the salient object detection phase, we apply this detector to two warped images that are to be stitched together. This process involves extracting the foreground objects from both the target image and the reference image, resulting in foreground object masks for each.

These masks are then combined to form a union mask of the warped images' foreground objects. The resulting mask provides semantic completeness information, which is subsequently fed into the Object Aware Seam Identification Module. This module uses the foreground object information to constrain and optimize the subsequent stitching process, ensuring that the final stitched image maintains high semantic integrity and visual coherence.

\subsection{Object Aware Seam Identification Module}

Following the acquisition of foreground object integrity information from the Salient Object Detector and feature maps from two Surface Feature Detectors, the Object Aware Seam Identification Module processes these inputs to generate seam masks and their corresponding soft-coded masks\cite{li2018perception}.

The Salient Object Detector, utilizing the SelfReformer method \cite{yun2022selfreformer}, provides semantic completeness for target and reference images. Concurrently, Surface Feature Detectors employ re-parameterized convolutions for downsampling and feature extraction, preserving differential information crucial for seam detection.

The module leverages these feature maps within a UNET-structured network\cite{ronneberger2015u}, optimized by the FastViT architecture\cite{vasufastvit2023}, to transform differential feature maps into seam feature maps. The seam generator then utilizes upsampling and re-parameterized convolutions, alongside a Sigmoid activation, to predict the seam mask. This mask is refined by eliminating invalid regions through multiplication with the aligned mask, resulting in a precise final seam mask.

By integrating semantic integrity and detailed feature maps, the Object Aware Seam Identification Module ensures accurate seam identification, enhancing the quality and coherence of the final stitched image.

\begin{algorithm}[h]
    \caption{Area based dynamic mask optimization}
    \begin{algorithmic}[1]
    \For{$i \leftarrow 1$ to $max\_epochs$}
        \State $M_1 \leftarrow O \odot L_1$ \Comment{Intersection of $O$ and $L_1$}
        \State $M_2 \leftarrow O \odot L_2$ \Comment{Intersection of $O$ and $L_2$}
        \State $A_{M_1} \leftarrow \sum_{i,j} M_1(i, j)$ \Comment{Area of $M_1$}
        \State $A_{M_2} \leftarrow \sum_{i,j} M_2(i, j)$ \Comment{Area of $M_2$}
        \If{$A_{M_1} > A_{M_2}$}
            \State $\mathcal{L}_{\text{comp}} \leftarrow \frac{1}{N} \sum_{i,j} (O(i, j) - M_1(i, j))^2$ 
        \Else
            \State $\mathcal{L}_{\text{comp}} \leftarrow \frac{1}{N} \sum_{i,j} (O(i, j) - M_2(i, j))^2$ 
        \EndIf
        \State $\mathcal{L}_{\text{excl}} \leftarrow \sum_{i,j} M_2(i, j)^2$ \Comment{Exclusivity loss}
        \State $\mathcal{L}_{\text{smooth}} \leftarrow \sum_{i,j} \left( \left( \frac{\partial L}{\partial x} \right)^2 + \left( \frac{\partial L}{\partial y} \right)^2 \right)$ \Comment{Smoothness loss}
        \State $\mathcal{L}_{\text{total}} \leftarrow \mathcal{L}_{\text{comp}} + \mathcal{L}_{\text{excl}} + \mathcal{L}_{\text{smooth}}$ 
        \Comment{Total loss}
        \State Take gradient descent step on $\mathcal{L}_{\text{total}}$ 
        \If{$\mathcal{L}_{\text{total}}$ converges}
            \State \textbf{break} \Comment{Stop training if loss converges}
        \EndIf
    \EndFor
    \end{algorithmic}
\end{algorithm}

\subsection{Image Stitching Module}

The Object Aware Seam Identification Module generates soft-coded masks \( L_1 \) and \( L_2 \) for the warped images \( I_1 \) and \( I_2 \), enabling seamless blending.

The final stitched image \( S \) is computed as:
\begin{equation}
S = L_1 \cdot I_1 + L_2 \cdot I_2
\label{eq:stitched_image}
\end{equation}

This method preserves semantic integrity and visual coherence by blending foreground and background smoothly, reducing artifacts.

\subsection{Dynamic Mask Optimization Based on Area}

To improve object preservation, we introduce a loss function that dynamically prioritizes the mask better covering the foreground object. 

\begin{figure}[t]
    \centerline{\includegraphics[width=1\linewidth]{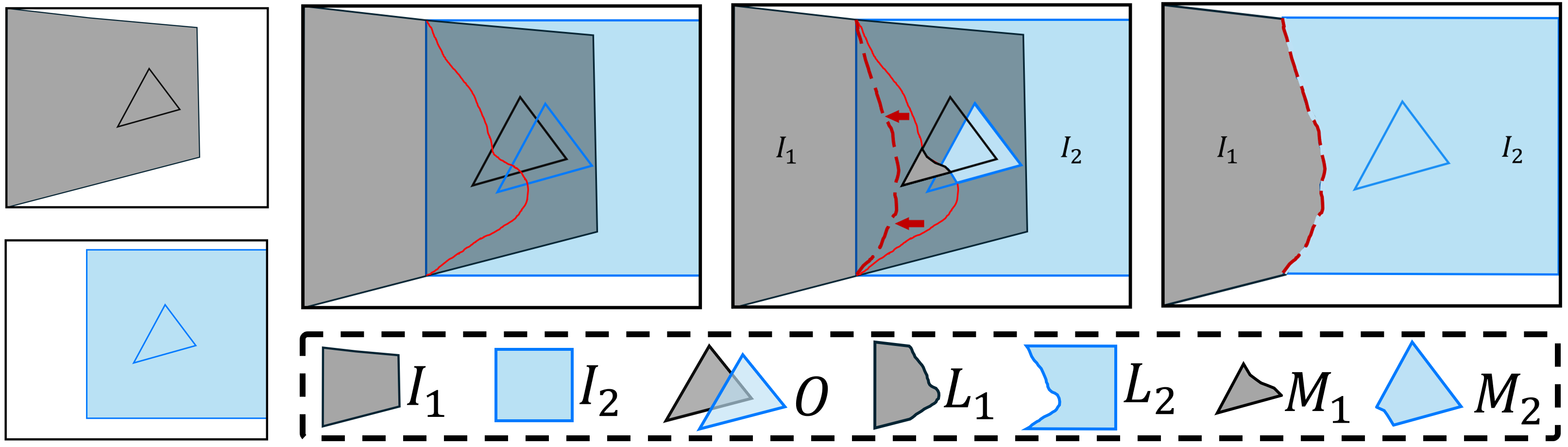}}
    \caption{\textbf{Symbol Definition:} Symbols and corresponding illustrations of dynamic mask optimization.}
    \label{fig2}
\end{figure}

As shown in Fig.~\ref{fig2}, \(I_1\) and \(I_2\) represent the warped \(image_1\) and \(image_2\), respectively. \(L_1\) and \(L_2\) are soft-coded masks \(\in [0,1]\), representing the masks corresponding to \(image_1\) and \(image_2\) learned by the neural network. \(O\) is a binary \(0/1\) mask representing the union of the foreground object masks extracted from \(image_1\) and \(image_2\). \(M_1\) and \(M_2\) are soft-coded masks \(\in [0,1]\), representing the intersection of \(O\) with \(L_1\) and \(L_2\), respectively. \( N \) is the total number of elements in \(\mathbf{O}\). \(A_{\mathbf{M}_1} \) and \(A_{\mathbf{M}_1} \) and \(A_{\mathbf{M}_2} \) represent the areas of \(M_1\) and \(M_2\), respectively.

The loss function \( \mathcal{L}_{\text{comp}} \) is defined as:

\begin{equation}
\mathcal{L}_{\text{comp}} = \frac{1}{N} \sum_{i,j} \left( \mathbf{O}(i,j) - \mathbf{M}_k(i,j) \right)^2
\label{eq:loss}
\end{equation}
where \(k = \arg\max \bigl(A_{\mathbf{M}_1}, A_{\mathbf{M}_2}\bigr)\).

This optimization dynamically selects the mask that best covers the object. It prioritizes regions with fewer background artifacts, reducing the severe background tearing caused by static mask selection. As shown in Fig.~\ref{fig2}, this dynamic process significantly improves image stitching quality, ensuring smoother transitions and fewer visual artifacts.

\subsubsection{Object Exclusivity Loss}
The object exclusivity loss is designed to minimize the overlap between \texttt{object\_mask} and \texttt{learned\_mask2}, ensuring that the second learned mask does not cover the object:
\begin{equation}
\mathcal{L}_{\text{excl}} = \sum_{i,j} \mathbf{M}_2(i,j)^2
\label{excel}
\end{equation}

\begin{figure*}[htbp]
    \centering
    \includegraphics[width=\textwidth]{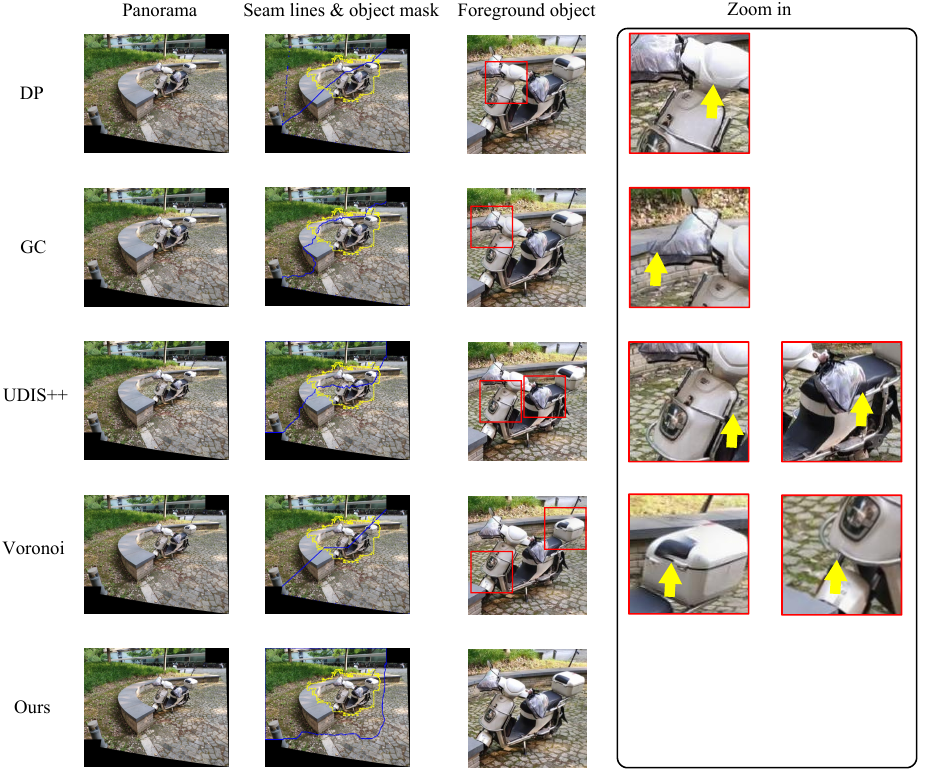}
    \caption{\textbf{Detailed comparative results :} Comparison of image stitching results using Dynamic Programming, Graph Cut, UDIS++, Voronoi, and our proposed method.}
    \label{fig:result_1}
\end{figure*}

\subsubsection{Smoothness Loss}
To ensure the smoothness of the learned masks, we introduce a smoothness loss term. This term penalizes abrupt changes in the mask values, promoting a smoother mask contour. Let \( \mathbf{L} \) be either \( \mathbf{L}_1 \) or \( \mathbf{L}_2 \):

\begin{equation}
    \mathcal{L}_{\text{smooth}} = \sum_{i,j} \left( \left( \frac{\partial \mathbf{L}}{\partial x} \right)^2 + \left( \frac{\partial \mathbf{L}}{\partial y} \right)^2 \right)
    \label{eq:smooth}
\end{equation}

\subsection{Combined Loss Function}

The total loss function is a combination of the object completeness loss, object exclusivity loss, and the smoothness loss. We denote this combined loss as the \textbf{Composite Coverage Loss}:
\begin{equation}
\mathcal{L}_{\text{total}} = \mathcal{L}_{\text{comp}} + \mathcal{L}_{\text{excl}} + \mathcal{L}_{\text{smooth}}
\label{loss_all}
\end{equation}

This method dynamically adjusts the focus of the learned masks based on the relative areas of their intersections with the object mask. By doing so, it ensures that the mask with the higher overlap is further encouraged to fully cover the object, thereby improving the accuracy of the object representation while maintaining exclusivity and smoothness of the mask contours.
\section{Experiments}

\subsection{Experiment Settings}
\textbf{Computational platform details:} Our experiments were conducted on a machine configured with Ubuntu 22.04, Intel i9-14900K CPU, NVIDIA 3090 GPU,and CUDA 11.

\textbf{Training time:} With a batch size of 1, our model occupied approximately 7.3GB of GPU memory and took about 22 minutes to train for 100 epochs.

\textbf{Inference time:} For a single input image of size 512x512 pixels, our model (33.55M parameters and 10.77 GFLOPs) required approximately 0.114 seconds for inference, utilizing roughly 3.2GB GPU memory.

\begin{figure*}[htbp]
    \centering
    \includegraphics[width=\textwidth]{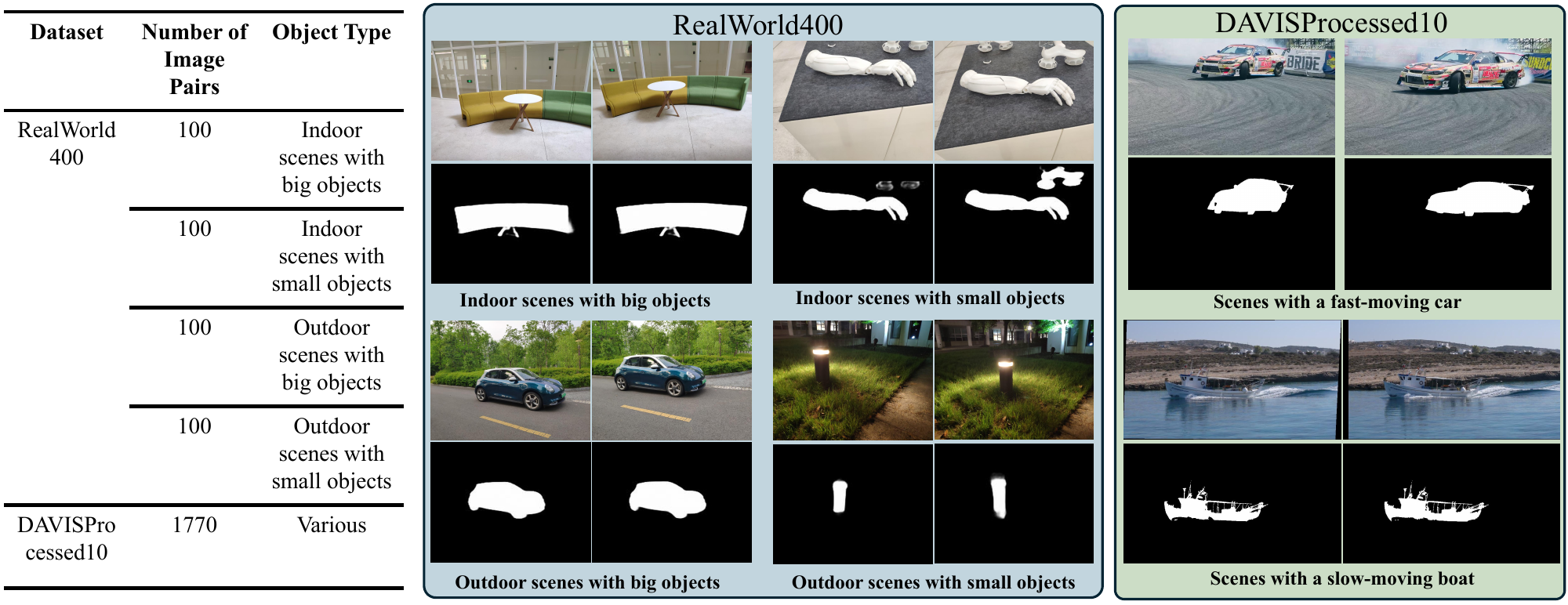}
    \caption{\textbf{Dataset overview:} An illustration and statistics of our dataset, covering diverse scenes.}
    \label{fig:dataset}
\end{figure*}

\subsection{Dataset}

\textbf{Training set:} To train our network, we utilized an unsupervised deep image stitching dataset, referred to as UDIS-D\cite{9472883}, which is derived from a variety of moving videos. Some of these videos are sourced from \cite{zhang2020content}, while others are captured independently. The UDIS-D dataset comprises 10,440 cases, encompassing diverse real-world scenes such as indoor, outdoor, night, dark, snow, and zooming conditions.

\textbf{Testing Set:} To address the challenge of seam lines intersecting foreground objects in image stitching tasks, traditional image stitching datasets and UDIS++\cite{Nie_Lin_Liao_Liu_Zhao_2023} may not adequately meet specific requirements. To better tackle this task, we designed and collected two specialized datasets.

First, we processed the original DAVIS\cite{perazzi2016benchmark} dataset by selecting the first and last frames from every ten-frame sequence, ensuring some displacement and angle variation. This yielded 1770 image pairs, designated as the DAVISProcessed10 test set. 

In addition, we constructed a paired real-world testing dataset, where each image pair contains foreground objects and covers a diverse range of scenes, including indoor, outdoor, daytime, and nighttime environments. This dataset consists of 400 image pairs in total. We employed SelfReformer\cite{yun2022selfreformer} to generate high-quality foreground object labels for each image, followed by manual post-processing to further refine and enhance their accuracy. The dataset is named RealWorld400, and its composition is illustrated in Figure \ref{fig:dataset}.

These two datasets are designed to encompass a variety of complex scenarios where seam lines intersect objects, thereby supporting the testing of deep learning models dedicated to this task. This dataset enhances model generalization and practicality, ensuring more accurate and natural stitching outcomes in real-world applications.

These datasets provide a robust foundation for developing and evaluating deep learning models aimed at improving image stitching accuracy, particularly in challenging scenarios involving object intersections. We believe this contribution holds significant value for the research community.

\subsection{Comparative Experiments}

To evaluate the performance of our proposed approach, we utilize aligned image pairs as input and benchmark against three established seam detection methods: dynamic programming (DP) \cite{duplaquet1998building}, the Voronoi-based approach (Voronoi) \cite{aurenhammer2000voronoi}, and Graph Cut \cite{nguyen2022efficient}. Additionally, we compare our method with the deep learning-based image stitching method, UDIS++.

\subsubsection{Metrics}

In this comparative experiment, we employed three commonly used image quality assessment metrics: Naturalness Image Quality Evaluator (NiQE)\cite{mittal2012making}, Blind/Referenceless Image Spatial Quality Evaluator (BRISQUE)\cite{mittal2012no}, and Perception-based Image Quality Evaluator (PIQE)\cite{moorthy2011blind}. These metrics are no-reference image quality assessment methods, enabling the direct evaluation of image quality without the need for a reference image. Specifically, NiQE assesses image quality by measuring the naturalness of the image, BRISQUE evaluates based on the spatial characteristics of the image, and PIQE considers human visual perception characteristics. Lower values for all these metrics indicate higher image quality. 

In addition to these general-purpose metrics, we also adopt the Perceptual Seam Quality (PSQ) measure\cite{zhang2025seam} to evaluate stitching performance along seamlines. PSQ quantifies local misalignments between overlapped patches and weights errors by visual saliency, emphasizing noticeable differences in salient regions. The score is normalized to [0, 1], where lower values indicate better seam quality and stitching performance.

Thus, we determine the effectiveness of different image stitching methods by comparing their scores across these three metrics.

\subsubsection{Quantitative Evaluation}
To validate the robustness of our proposed method, Table \ref{tab:comparison_methods} presents the quantitative evaluation results of different image stitching methods across three datasets. UDIS-D is a publicly recognized and widely accepted image stitching dataset. Additionally, to better assess whether our method optimizes the integrity of foreground objects, we conducted evaluations on the DAVISProcessed10 and RealWorld400 datasets.

We conducted a comparative analysis against three established traditional seam detection methods : Dynamic Programming (DP) \cite{duplaquet1998building}, the Voronoi-based approach (Voronoi) \cite{aurenhammer2000voronoi}, and Graph Cut (GC) \cite{nguyen2022efficient} as well as the deep learning-based image stitching methods like UDIS++\cite{Nie_Lin_Liao_Liu_Zhao_2023}, Recdiffusion\cite{zhou2024recdiffusion}, SRStitcher\cite{xie2024reconstructing} and TRIS\cite{jiang2024towards}.

Our method outperforms existing stitching methods across all datasets and most metrics, demonstrating superior performance and stability in the image stitching task. These results indicate that our method excels not only on synthetic datasets but also exhibits strong adaptability and robustness in real-world scenarios.

\begin{table*}[ht]
\centering

\resizebox{\textwidth}{!}{
\begin{tabular}{ccccccccccccc}
\toprule
\multirow{2}{*}{Method} & \multicolumn{4}{c}{UDIS-D} & \multicolumn{4}{c}{DAVISProcessed10} & \multicolumn{4}{c}{RealWorld400} \\
\cmidrule(lr){2-5} \cmidrule(lr){6-9} \cmidrule(lr){10-13}
& Niqe $\downarrow$ & BRISQUE $\downarrow$ & PIQE $\downarrow$ & PSQ $\downarrow$ & Niqe $\downarrow$ & BRISQUE $\downarrow$ & PIQE $\downarrow$ & PSQ $\downarrow$ & Niqe $\downarrow$ & BRISQUE $\downarrow$ & PIQE $\downarrow$ & PSQ $\downarrow$ \\
\midrule
GC & 6.035 & 41.03 & 25.32 & 0.32 & 4.837 & 22.73 & 15.65 & 0.28 & 5.19 & 29.35 & 13.88 & 0.37 \\
DP & 5.995 & 40.84 & 25.29 & 0.35 & 4.816 & 22.35 & 15.73 & 0.29 & 5.181 & 29.39 & 14.09 & 0.27 \\
Voronoi & 6.030 & 41.01 & 25.31 & 0.38 & 4.809 & 22.61 & 15.81 & 0.33 & 5.195 & 29.42 & 13.93 & 0.29\\

TRIS & 4.620 & 37.96 & 23.17 & 0.14 & \underline{3.381} & 20.93 & 15.08 & \underline{0.13} & 3.862 & 26.72 & 12.43 & 0.16 \\
SRStitcher & 4.830 & 39.62 & \underline{22.78} & \underline{0.12} & 3.452 & 21.32 & 15.27 & 0.18 & \underline{3.284} & 27.16 & \underline{11.83} & 0.13 \\
Recdiffusion & 4.450 & 38.56 & 23.12 & 0.15 & 3.429 & 20.86 & 15.62 & 0.17 & 3.373 & 26.83 & 12.15 & 0.17 \\

UDIS++ & \underline{4.209} & \underline{37.84} & 22.97 & 0.17 & 3.448 & \textbf{20.14} & \textbf{13.75} & 0.14 & 3.312 & \underline{26.37} & 11.98 & \underline{0.11} \\
Ours & \textbf{4.169} & \textbf{37.57} & \textbf{22.60} & \textbf{0.10} & \textbf{3.296} & \underline{20.77} & \underline{15.03} & \textbf{0.11} & \textbf{3.188} & \textbf{26.48} & \textbf{11.75} & \textbf{0.09} \\
\bottomrule
\end{tabular}
}
\caption{\textbf{Quantitative evaluation: }Our method consistently achieves top performance across almost all datasets.}
\label{tab:comparison_methods}
\end{table*}

\subsubsection{Qualitative Evaluation}

\begin{figure*}[htbp]
    \centering
    \includegraphics[width=\textwidth]{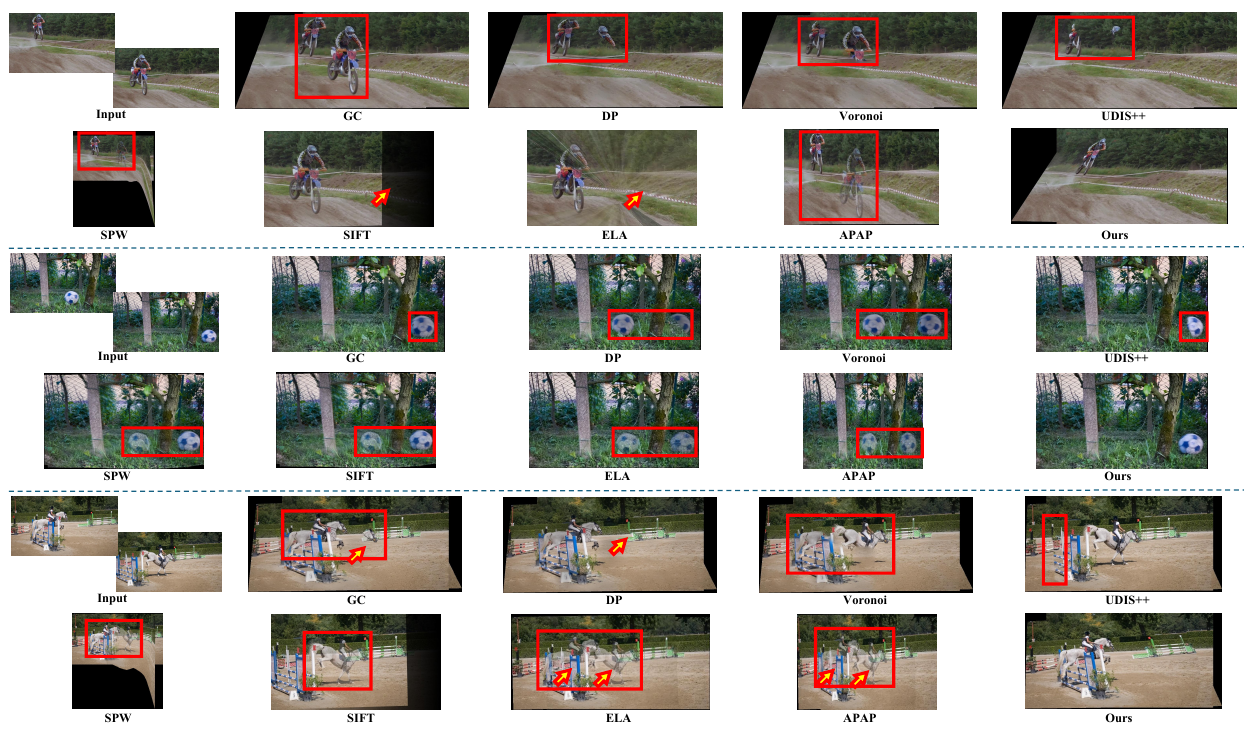}
    \caption{\textbf{Comparative results on real-world datasets : }Foreground object stitching failures are highlighted with red boxes, and severe misalignment issues are indicated with yellow arrows.}
    \label{fig:result_2}
\end{figure*}

To visually demonstrate the effectiveness of our proposed method, we provide a visualization of the above methods, as shown in Figure \ref{fig:result_1} and Figure \ref{fig:result_2}.

\begin{table}[h]
\centering
\label{tab:user_study_results}
\begin{tabular}{lccc}
\hline
\textbf{Method} & \textbf{Coherence} & \textbf{Integrity} & \textbf{Quality} \\ \hline
DP              & 3.6                & 3.5                & 3.4              \\
GC              & 3.8                & 3.7                & 3.6              \\
TRIS          & 4.0                & 4.3                & 4.6              \\
SRStitcher          & 4.4                & 4.6                & 4.7              \\
Recdiffusion          & 4.5                & 4.7                & 4.2              \\
UDIS++          & 4.2                & 4.1                & 4.1              \\
\textbf{Our Method} & \textbf{4.8}   & \textbf{4.9}       & \textbf{4.8}     \\ \hline
\end{tabular}
\caption{\textbf{User study results on RealWorld400 dataset : }The results of the user study on the RealWorld400 dataset.}
\label{tab:user_study_results}
\end{table}

The results of this analysis are illustrated in Figure \ref{fig:result_1}, which provides a detailed comparison of the stitched images produced by each method.Figure \ref{fig:result_1} shows the final stitched panoramas, seam lines with object masks, foreground objects, and zoomed-in views of critical seam intersections. The yellow arrows in the zoomed-in views indicate notable foreground object discontinuities, breaks at the seams, and artifacts.Our proposed method demonstrates superior performance in preserving the integrity of foreground objects and minimizing visible artifacts at seam intersections, outperforming the other methods.

Furthermore, to more comprehensively demonstrate the superiority and robustness of our approach, we compared it with non-seam detection methods such as SPW\cite{liao2019single}, SIFT\cite{lowe2004distinctive}, APAP\cite{Zaragoza_Chin_Tran_Brown_Suter_2014} and ELA\cite{Li_Wang_Lai_Zhai_Zhang_2018}. As illustrated in Figure \ref{fig:result_2}, these non-seam methods exhibit significant ghosting and misalignment issues. In contrast, our method consistently achieves the highest quality results, free from these common artifacts, thereby affirming its superior performance.

\begin{figure}[htbp]
    \centering
    \includegraphics[width=1\linewidth]{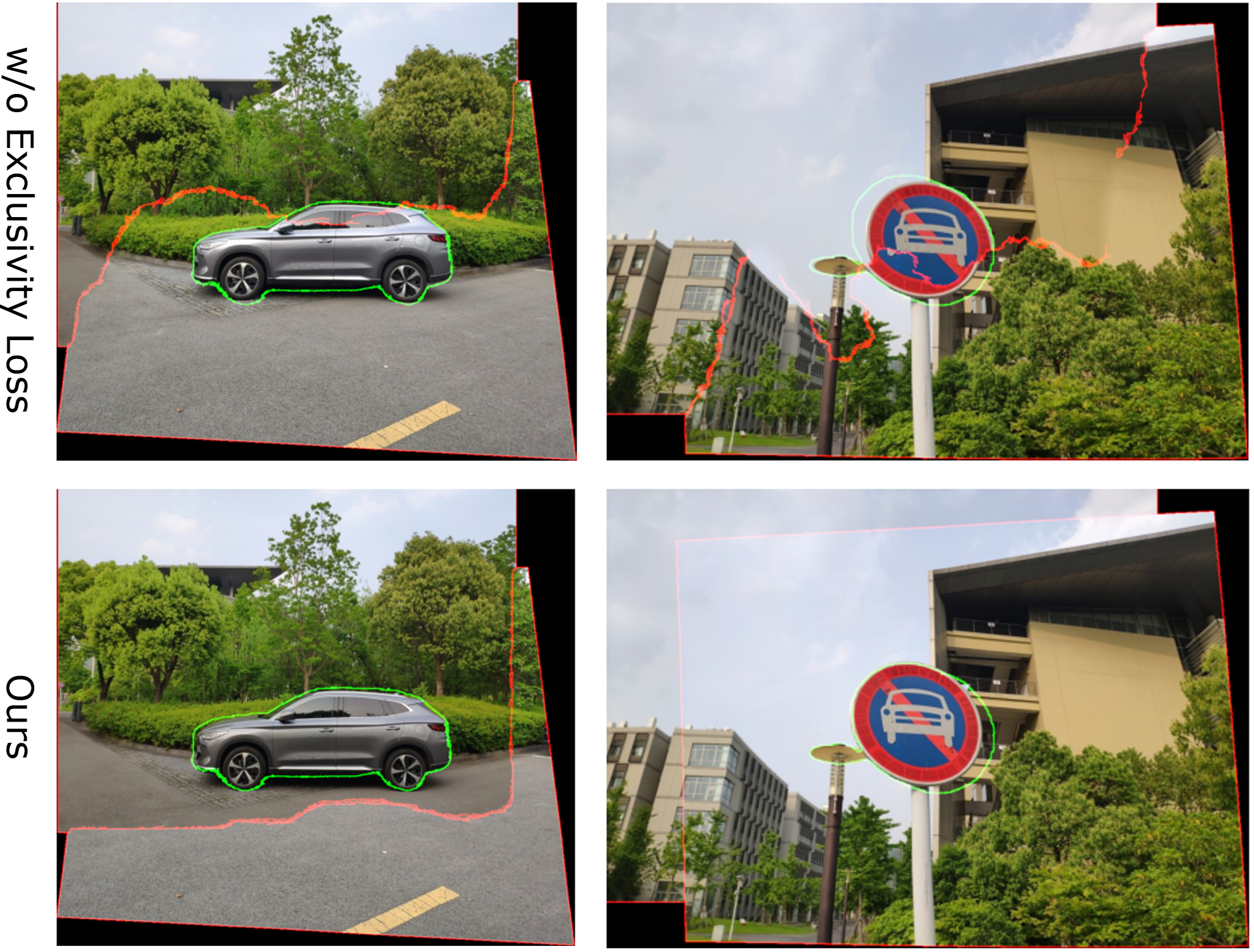}
    \caption{\textbf{Object exclusivity loss : } Red indicates the stitching line, and the green box denotes the detected foreground object.}
    \label{fig:al_loss}
\end{figure}

\subsubsection{User Study}

To evaluate the subjective quality of our proposed image stitching method, we conducted a user study with 50 participants using the RealWorld400 dataset. Each participant was randomly assigned 40 image pairs, ensuring that each pair was reviewed by at least five different participants. The image pairs included one image stitched using our method and another using a comparison method (DP, GC, or UDIS++).

Participants rated each pair based on visual coherence, foreground object integrity, and overall image quality on a scale from 1 to 5. The results, summarized in Table \ref{tab:user_study_results}, indicate that our method consistently received higher ratings across all criteria.

These findings corroborate the quantitative assessments and demonstrate the superiority of our method in preserving visual coherence and foreground object integrity, enhancing the overall image quality.

\begin{table}[h!]
\centering
\begin{tabular}{c c c}
\hline
\textbf{Algorithms} & \textbf{\#Params(M)} & \textbf{GFLOPs} \\
\hline
UDIS++ & 33.56 & 80.46 \\
\textbf{Ours} & \textbf{33.55} & \textbf{10.76} \\
\hline
\end{tabular}
\caption{Comparison of parameters (in millions) and GFLOPs between UDIS++ and our model for 512x512 input images.}
\label{tab:ablation_1}
\end{table}

\subsection{Ablation Study}

We conducted an ablation study to evaluate the impact of our Object Aware Seam Identification Module, comparing the baseline model (UDIS++) with our enhanced model. Both models have similar parameter counts, around 33.56 million, but our model reduces the computational cost from 80.46 GFLOPs to 10.77 GFLOPs, as shown in Table \ref{tab:ablation_1}. The input image size for this evaluation was 512x512.

\vspace{1em}
\begin{table}[htbp]
    \centering
    \label{tab:ablation_study}
    \renewcommand{\arraystretch}{1.2}
    \setlength{\tabcolsep}{4pt} 
    \resizebox{\linewidth}{!}{
    \begin{tabular}{lccccc}
        \toprule
        \textbf{Method} & \textbf{PIQE} & \textbf{BRI.} & \textbf{NIQE} & \textbf{Failure Cases} & \textbf{Rate $\uparrow$} \\
        \midrule
        w/o Dynamic Mask Opt. & 14.29 & 20.43 & 3.352 & 64 & 84\% \\
        w/o Exclusivity Loss & 13.75 & 20.14 & 3.448 & 132 & 67\% \\
        Ours & 15.03 & 20.77 & 3.296 & 52 & 87\% \\
        \bottomrule
    \end{tabular}}
    \caption{\textbf{Ablation study results : }The results of the ablation study on the ForegroundStitch-Part I dataset.}
    \label{tab:ablation_study}
\end{table}

These results demonstrate that our Object Aware Seam Identification Module significantly reduces computational cost while maintaining a similar parameter count, highlighting its efficiency and effectiveness.

To validate the effectiveness of our proposed dynamic mask optimization and object exclusivity loss, we conduct ablation studies on ForegroundStitch-Part I. As shown in Table \ref{tab:ablation_study}, these strategies enhance the integrity of foreground objects in the stitched images while maintaining acceptable overall image quality. In particular, Figure \ref{fig:al_loss} illustrates the role of the object exclusivity loss: when foreground objects are present in the images to be stitched, this loss effectively prevents seam lines from passing through salient objects, thereby avoiding noticeable inconsistencies in perspective within the foreground regions.

\begin{figure}[htbp]
    \centering
    \includegraphics[width=1\linewidth]{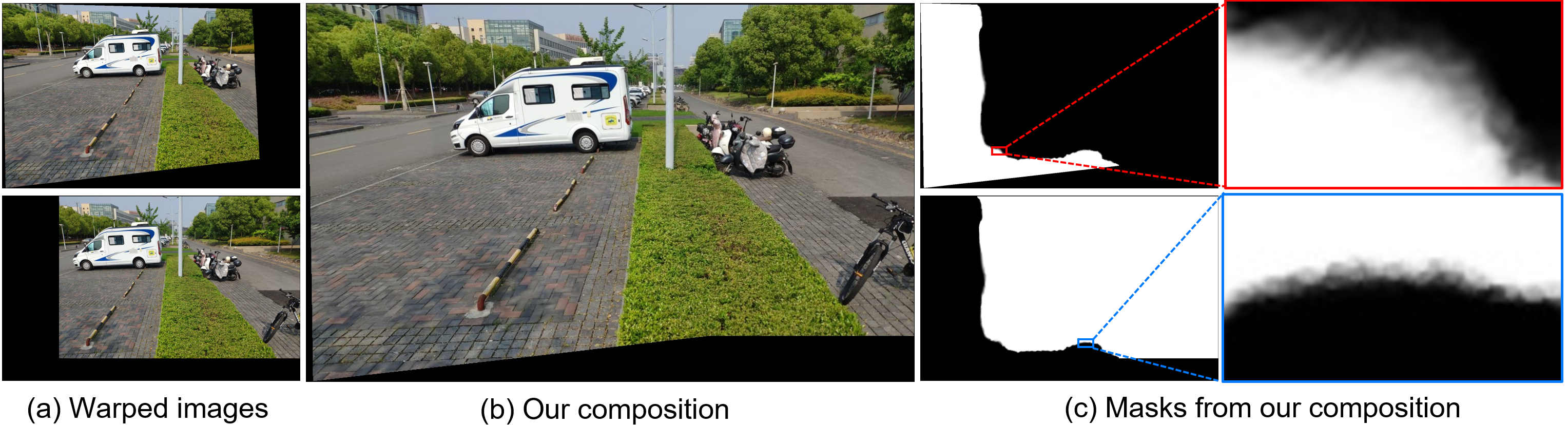}
    \caption{\textbf{Dynamic mask optimization : }Preserves object integrity by assigning pixels to a single source image, while soft-coded masks ensure smooth seam transitions.}
    \label{fig:al_dynamic_mask}
\end{figure}

Furthermore, Figure~\ref{fig:al_dynamic_mask} demonstrates the effect of dynamic mask optimization. 
This strategy ensures that all pixels of a foreground object originate from the same source image 
rather than being split between two images, thus preserving object integrity. 
In addition, as shown in Figure~\ref{fig:al_dynamic_mask} (c) , 
we adopt soft-coded masks with values in $[0,1]$ instead of hard binary masks $\{0,1\}$, 
which guarantees smoother transitions along the seam boundaries 
and leads to more natural stitching results.

\section{Application}  

Panoramic photography stitches multiple images to create wide-angle views but often suffers from ghosting and misalignment, especially with moving objects. These artifacts result from discontinuities introduced when stitching frames with dynamic elements, as shown in Figure \ref{fig:panorama} (a). Traditional smartphone panoramas, such as the iPhone 14 Pro example in Figure \ref{fig:panorama} (c), often display duplication and misalignment of moving objects, leading to distorted or fragmented panoramas.

Our method, illustrated in Figure \ref{fig:panorama} (b) and the close-up view in (d), addresses these issues by preserving the integrity of foreground objects and reducing duplication. Although it does not guarantee a single instance of each moving object, it significantly improves object continuity and enhances overall image quality. This makes our approach ideal for capturing dynamic scenes, ensuring accurate alignment and a seamless panoramic experience.

\section{Limitations}
Despite our method's effectiveness in preventing the foreground object from being fragmented, it comes at the cost of introducing discontinuities at the edges of the two images. As illustrated in the Figure 8, our approach successfully maintains the integrity of the foreground object. However, the boundary regions between the images exhibit noticeable segmentation artifacts, underscoring a significant limitation of our current implementation.

\begin{figure}[t]
    \centering
    \includegraphics[width=1\linewidth]{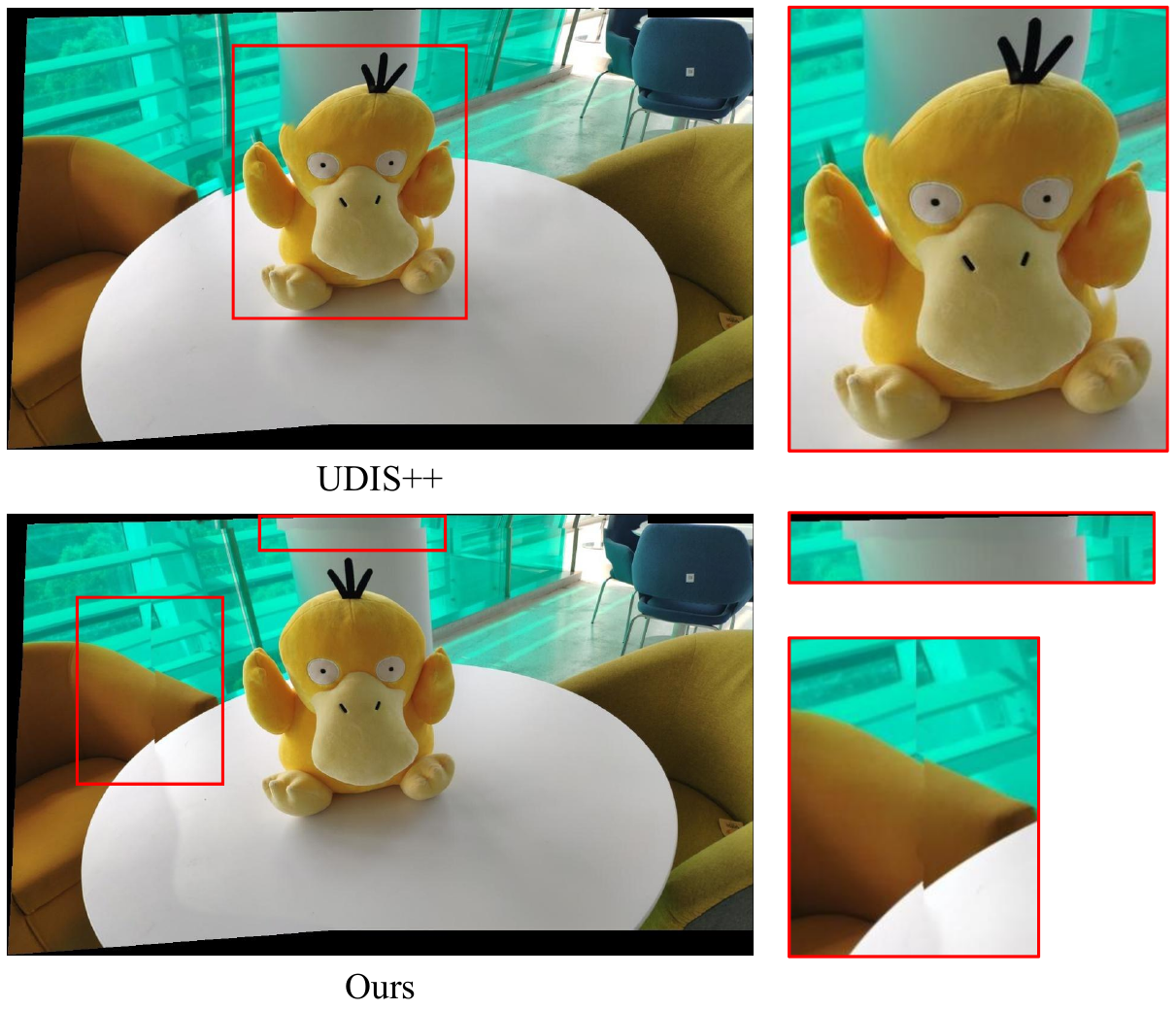}
    \caption{\textbf{Comparison of image stitching results between UDIS++[18] and our method : }The edge discontuities are highlighted.}
    \label{fig:limitation}
\end{figure}

\begin{figure}[t]
    \centering
    \includegraphics[width=1\linewidth]{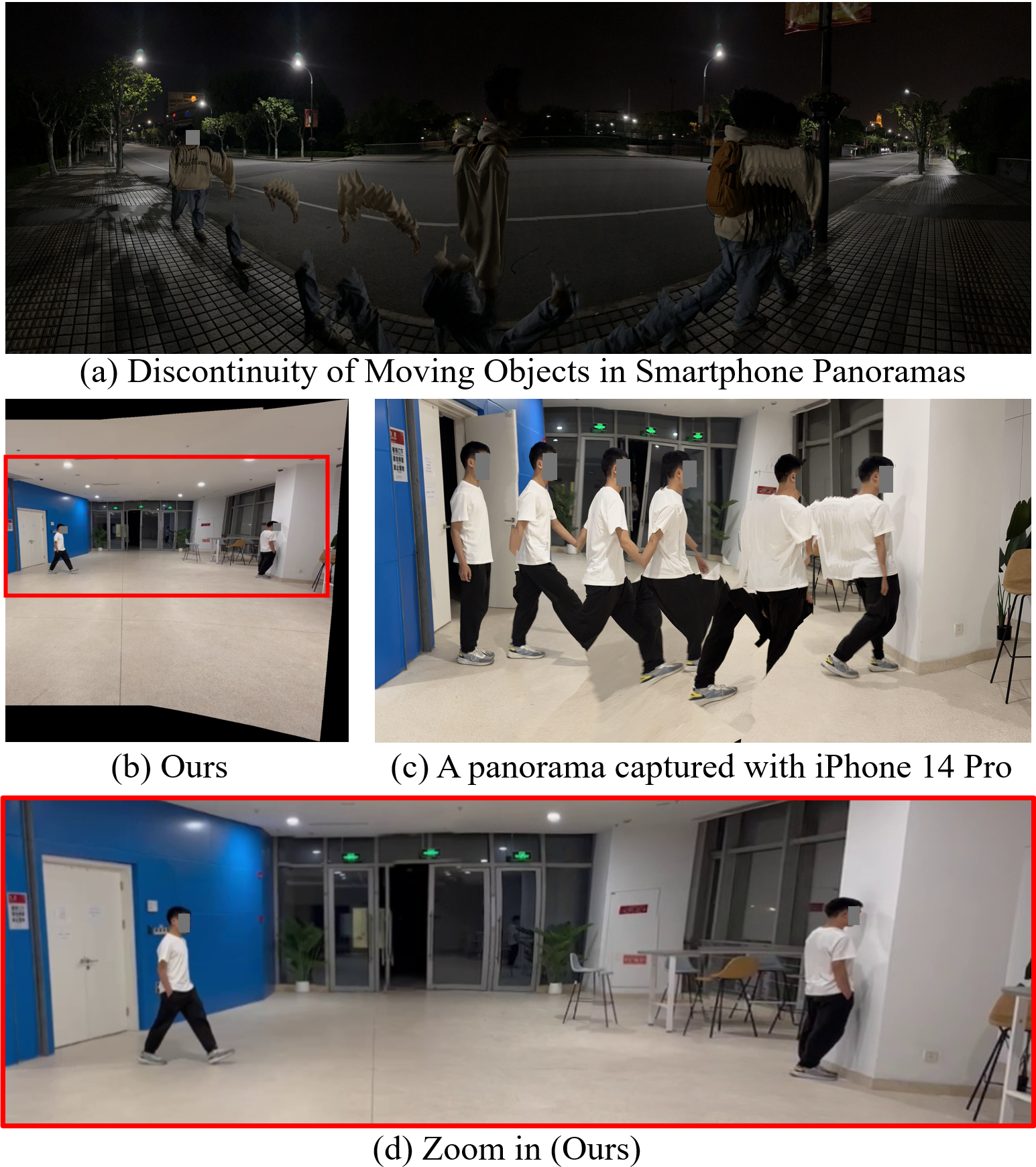}
    \caption{
    \textbf{Comparison with existing methods : }Our method(a) avoids duplication and misalignment, outperforming the iPhone 14 Pro(b). The zoomed-in section(c) highlights this improvement.}
    \label{fig:panorama}
\end{figure}

\section{Conclusion}

In this paper, we introduced an advanced deep learning-based framework for image stitching that prioritized semantic integrity and visual coherence. Our approach leveraged semantic priors of foreground objects to avoid seam lines intersecting critical areas, ensuring smooth transitions and enhanced image quality. We proposed a novel loss function designed to preserve the semantic attributes of salient objects, which significantly improved the realism and overall aesthetic of the stitched images. Additionally, we constructed specialized real-world datasets to thoroughly evaluate our method, demonstrating its robustness and practical applicability.

The experimental results indicated that our method substantially outperformed traditional and contemporary techniques in both qualitative and quantitative assessments. The comprehensive ablation study validated the efficiency and effectiveness of our Object Aware Seam Identification Module, highlighting its significant reduction in computational cost without compromising performance.

Our user study provided further validation, with participants consistently rating our method higher in visual coherence, foreground object integrity, and overall image quality. These findings corroborate our quantitative assessments, reinforcing the superiority of our approach in real-world applications.

\bibliographystyle{spmpsci}
\bibliography{ref} 

\end{document}